%% file: main.tex
\pdfoutput=1
\documentclass[11pt]{article}

\usepackage[final]{acl}
\def\shownotes{1}
\usepackage{macro}

\usepackage{times}
\usepackage{latexsym}
\usepackage{makecell}
\usepackage{balance}

\usepackage[T1]{fontenc}

\usepackage[utf8]{inputenc}

\usepackage{microtype}

\usepackage{inconsolata}

\usepackage{graphicx}
\usepackage{balance}
\usepackage{caption}
\usepackage{subcaption}
\usepackage{booktabs}
\usepackage{algorithm,algorithmicx,algpseudocode}
\usepackage{enumitem}
\usepackage{multirow}
\makeatletter
\def\thanksnosymbol#1{\protected@xdef\@thanks{\@thanks
        \protect\footnotetext{#1}}}
\makeatother
\title{Scalable Fine-tuning from Multiple Data Sources:\\A First-Order Approximation Approach}
\author{Dongyue Li\textsuperscript{\dag*}\thanksnosymbol{\textsuperscript{*}Equal Contribution.  Emails: {\{li.dongyu, zhang.zini, ho.zhang\}@northeastern.edu} and {wangluxy@umich.edu}.} \ \ \ \ Ziniu Zhang\textsuperscript{\dag*} \ \ \ \ 
Lu Wang\textsuperscript{\ddag} \ \ \ \ Hongyang R. Zhang\textsuperscript{\dag} \\
        \textsuperscript{\dag}Northeastern University, Boston, MA\\
        \textsuperscript{\ddag}University of Michigan, Ann Arbor, MI
}

\begin{document}
\maketitle

\input{content}

\input{experiment}
\input{related}
\bibliography{ref}

\clearpage
\appendix
\input{appendix}

\end{document}

%% file: content.tex
\begin{abstract}
We study the problem of fine-tuning a language model (LM) for a target task by optimally using the information from $n$ auxiliary tasks. This problem has broad applications in NLP, such as targeted instruction tuning and data selection in chain-of-thought fine-tuning. The key challenge of this problem is that not all auxiliary tasks are beneficial in improving the performance of the target task. Thus, selecting the right subset of auxiliary tasks is crucial. Conventional subset selection methods, such as forward and backward stepwise selection, are unsuitable for LM fine-tuning because they require repeated training on subsets of auxiliary tasks. This paper introduces a new algorithm for estimating model fine-tuning performance without requiring repeated training. Our algorithm first performs multitask training using data from all tasks to obtain a meta initialization. Then, we approximate the model fine-tuning loss of a subset using functional values and gradients from the meta initialization. Empirically, we find that this gradient-based approximation holds with remarkable accuracy for twelve transformer-based LMs. Thus, we can now estimate fine-tuning performances on CPUs within a few seconds. Finally, we fine-tune the pretrained base model once on the selected subset of tasks. We conduct extensive experiments to validate this approach, delivering a speedup of $30\times$ over conventional subset selection while incurring only $1\%$ error of the true fine-tuning performances. In downstream evaluations involving both instruction tuning and chain-of-thought fine-tuning, this loss-based selection approach improves over prior gradient or representation similarity-based methods for subset selection by up to $3.8\%$.
\end{abstract}

\section{Introduction}

Fine-tuning a language model (LM) has emerged as an effective approach for knowledge transfer with text data.
As the scale of LMs continues to grow, efficient and scalable fine-tuning methods are in high demand.
For instance, parameter-efficient fine-tuning with adapters or low-rank parameterization can significantly reduce the memory usage of fine-tuning on a large model  \cite{houlsby2019parameter,pfeiffer2020mad,hu2021lora}.
In many applications, besides solving a target task of interest, one can also access several related data sources that can be used for data augmentation. Examples include multilingual systems such as neural machine translation \cite{neubig2018rapid}, parsing \cite{ustun2020udapter}, data selection \cite{xie2023data}, and targeted instruction tuning \cite{xia2024less}.
A fundamental issue in these scenarios is selecting the beneficial data sources amongst all the available data sources, which can be formulated as a \emph{subset selection} problem.
In this paper, we develop efficient algorithms for subset selection in LM fine-tuning given $n$ data sources, which can scale up to handle large numbers of $n$.

\begin{figure*}[t!]
    \centering
    \includegraphics[width=0.99\textwidth]{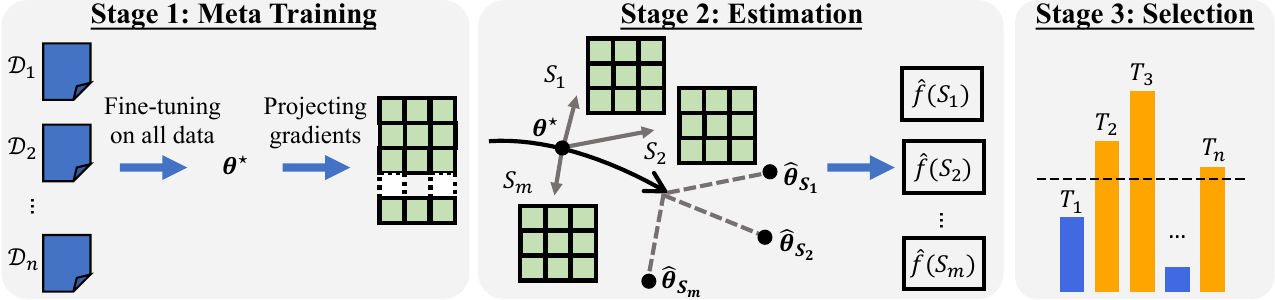}
    \caption{An overview of our approach: \textbf{(1)} Perform multitask training on a base LM using all the samples, leading to a meta-initialization $\theta^\star\in\real^p$. We store (randomly) projected gradients of $\theta^{\star}$ for every sample.
    \textbf{(2)} Estimate model fine-tuning performances on a list of task subsets using projected gradients as features in logistic regression.
    \textbf{(3)} Using the estimated results (denoted as $\hat f(S_1), \dots, \hat f(S_m)$), compute a score $T_i$ for each auxiliary task for $i = 1, 2, \dots, n$, which indicates its relevance to the target task.
    Finally, select a subset (depicted in blue color) using a threshold score, and fine-tune a pretrained LM by combining the training data of the selected tasks together.}\label{fig_pipeline}
\end{figure*}

Classical subset selection methods, such as forward and backward stepwise selection, have been very effective for feature selection in regression analysis \cite{hastie2009elements}. 
However, these classical methods are unsuitable for LM fine-tuning because they require repeated training of the base LM on many subsets of tasks, which is not feasible for large $n$.
More recently, data selection methods have been proposed based on selecting influential samples by comparing gradients \cite{xia2024less} or feature representation during training \cite{ivison2023data}.
These methods depend on the training procedure, leading to noisy outcomes for approximating true fine-tuning performances.
In summary, existing subset selection methods require expensive computation or rely on noisy measurements of task-relatedness \citep{vu2020exploring,vu2021spot,zhang2023survey}.

This paper presents a new approach to scale up subset selection by quickly estimating model fine-tuning losses without running the fine-tuning procedure.
The key idea is to first perform multitask training on samples from all tasks, leading to a \emph{meta initialization} $\theta^{\star}$.
From $\theta^{\star}$, we compute all the samples' functional values and gradients, which can be used during inference.
Second, given a subset of auxiliary tasks, we apply Taylor's expansion on the loss and approximate the loss value with the function values and the gradients of the samples in the subset.
After applying this approximation, we will estimate the fine-tuning losses directly with logistic regression without conducting any actual fine-tuning.
Notice that this computation can be done on CPUs, and we use random projections to reduce the dimension of the regression down to a few hundred.
Using this procedure, we can estimate the fine-tuning performance for each subset in just a few seconds, which is significantly faster than fine-tuning the full model.
Importantly, the accuracy of this approach hinges on the quality of the first-order approximation.
We find that across twelve transformer-based LMs, including Llama-3-8B, the approximation error of the gradient-based expansion is at the order of $10^{-5}$ to $10^{-3}$, in regions close to $\theta^{\star}$.
Based on this estimation, we can apply a classical subset selection method, such as forward selection, to estimate the loss values of every subset encountered during the procedure.
Lastly, we fine-tune a pretrained model by combining data from the selected subset of tasks.
In summary, our approach has three stages. See Figure \ref{fig_pipeline} for an illustration of our overall approach.

We conduct extensive experiments to validate our approach using four benchmarks.
We show that \algo{} can accurately approximate fine-tuned model losses within \textbf{1}\% errors across various LMs. Meanwhile, \algo{} can accelerate forward selection with \textbf{30}$\times$ fewer FLOPs and \textbf{25}$\times$ less GPU hours by circumventing full model fine-tuning. \algo{} also speeds up random ensemble with \textbf{44}$\times$ fewer FLOPs and \textbf{46}$\times$ less GPU hours.
We consider targeted instruction tuning and chain-of-thought fine-tuning for downstream evaluations of \algo{}. 
Our approach matches the accuracy of classical subset selection methods with full fine-tuning while incurring only \textbf{0.5}\% of the computational cost. On the ToxiGen and TruthfulQA benchmarks, \algo{} outperforms existing selection methods that rely on gradient or representation measurements \cite{xie2023data,xia2024less} by \textbf{3.8}\% on average. On StrategyQA and CommonsenseQA datasets, the loss-based selection of \algo{} improves chain-of-thought reasoning accuracy by \textbf{2.4}\% over gradient-based and feature-based selection methods.

In conclusion, this paper presents a novel algorithm for subset selection in LM fine-tuning, utilizing multiple data sources.
The algorithm can quickly estimate model fine-tuning performances without performing fine-tuning on each subset.
We identify a linearization property of LMs (after meta-training), which is empirically verified on twelve transformer-based LMs. This empirical finding may be relevant for future developments of fast inference methods of LLMs.
Our estimation algorithm unlocks classical subset selection methods, such as forward \& backward selection, to LMs.
The extensive experiments show that our approach can deliver significant speedups without losing downstream accuracy. In particular, our loss-based approach can outperform existing gradient-based or feature-based selection methods for instruction tuning and chain-of-thought fine-tuning.
The code repository and detailed instructions for reproducing our findings are available at \url{https://github.com/VirtuosoResearch/Scalable-finetuning}.

\section{Problem Formulation}\label{sec_motivation}

Consider fine-tuning a language model to solve a target task.
Suppose we also have access to $n$ source tasks.
Because the information from different data sources may conflict with each other or hurt the target performance \cite{wu2019understanding}, we would like to select a subset of the $n$ tasks.
This often happens in transfer learning, for instance, working with low-resource neural machine translation \cite{zoph2016transfer}.
Because there are $2^n$ possible choices of subsets, performing the best subset selection is very costly.

In this paper, we develop a significantly faster approach for subset selection that is well-suited for LLMs.
Concretely, given a base fine-tuning procedure on top of an LM (such as LoRA or QLoRA \cite{dettmers2024qlora}), let $f(S)$ be the model fine-tuning loss on the target task trained together with a given subset of auxiliary tasks $S\subseteq\set{1,2, \dots, n}$.
A lower value of $f$ indicates that $S$ is more relevant to the target task.
Best subset selection corresponds to finding a subset that minimizes $f(S)$.

\subsection{Task-relatedness measures}

\textbf{Experimental setup:} We will present a case study using the Alpaca dataset \cite{alpaca}, which involves $52,000$ instruction-following data generated by GPT-3, each containing an instruction as a task description and the corresponding task input and output.
The dataset can be pre-processed into $38$ tasks based on the instruction types \cite{wang2023self}. Each task is identified by the verb in the instruction, such as ``edit'' and ``describe.''
We randomly sample $10\%$ as the validation set and use the rest of $90\%$ samples as the training set.
We use TinyLlama-1.1B as the base LM and LoRA as the base protocol \cite{hu2021lora}.

$\diamondsuit$ First, we find the existence of negative transfers, which remains consistent with prior studies of multitask learning in NLP \cite{vu2020exploring,vu2021spot}.
Specifically, we compare (A1) fine-tuning the base LM on each task and (A2) fine-tuning the base LM on each task with \emph{one of} the remaining $37$ tasks as auxiliary tasks.
For $18 / 38$ tasks, A1 performs better than A2, indicating negative transfers from auxiliary tasks to the target task.

$\diamondsuit$ Second, we also find that $f(S)$ is not monotone, in the sense that adding an extra task to $S$ may not necessarily reduce the value of $f$ (although it does more data).
We pick a target task $t$ corresponding to ``arrange'' in Alpaca and find that $f(S)$ starts to increase after we add more than one task to $S$, which confirms our hypothesis. %
This suggests that directly optimizing $f$ over the space of subsets remains challenging because $f$ is complex.

$\diamondsuit$ Third, we also find that methods that utilize model gradients \cite{fifty2021efficiently} or representation similarity \cite{wu2019understanding,vu2020exploring} to select tasks do not correlate well with actual values of $f$.
For every pair of tasks $i$ and $j$, we evaluate the correlation between the cosine similarity of the averaged gradient of tasks $i,j$, and $f(\set{i, j})$.
We find that this is less than $0.2$.

\subsection{Problem statement}\label{sec_problem_statement}

Having described the nuances of $f$ and the potential complexity in optimizing $f$ directly (see also \citet{zhang2023survey} for further discussion on this topic), we now pose the following research question:
\emph{Given a list of subsets $S_1, \dots, S_m  \subseteq \set{1, \ldots, n}$, can we efficiently estimate $f(S_1), \dots, f(S_m)$, without fine-tuning the model on each possible subset?}
Next, we give two examples to illustrate the choices of these $m$ subsets.
\begin{example}[Forward stepwise selection]\label{ex_fs}
    In forward selection \cite{hastie2009elements}, one starts with an empty subset $S_1 = \set{}$. Then, enumerate through all singleton sets, $f(\set{1}), \ldots, f(\set{n})$, and pick the best one. Suppose $i_1$ is chosen. Then, evaluate $f(\set{i_1,1}), \dots, f(\set{i_1,n})$ except when $i_1$ is repeated, and pick the best one. Until $f$ peaks. %
\end{example}

\begin{example}[Random ensemble]\label{ex_re}
Random ensembling is a highly effective approach for data attribution \cite{ilyas2022datamodels}, which can also be used for subset selection in multitask learning \cite{li2023identification,li2023boosting,li2024scalable}.
We choose $m$ random subsets from $\set{1, 2, \dots, n}$ with a fixed size $\alpha$.
For example, if $\alpha = 2$, $f$ measures pairwise affinity scores between two tasks \cite{fifty2021efficiently}.
If $\alpha > 2$, $f(S_i)$ measures higher-order affinity scores between multiple tasks \cite{li2023boosting}.
\end{example}
If we can quickly estimate the values of $f(S_1), f(S_2), \dots, f(S_m)$, then we can still apply the above subset selection methods.
In the next section, we present our estimation approach to obtain these loss values.

\section{Our Approach}\label{sec_method}

\begin{table*}[ht!]
\centering
\caption{We empirically find that the \emph{{first-order approximation holds with very high accuracy within 0.25\% relative distance to the meta initialization $\theta^{\star}$}}. As shown in the table below, the RRSS is at the order of $10^{-5}$-$10^{-3}$ when $X$ is between $0.05\%$ to $0.25\%$ distance of $\theta^{\star}$, where $\theta^{\star}$ includes all model weights in the initialization. We attribute this behavior to the highly overparameterized nature of LLMs. As a remark, we use LoRA as the base fine-tuning behavior; thus, the fine-tuned distance is small because of using LoRA. We report the average over $50$ random task subsets to ensure statistical significance.
The reference for each LM can be found in Appendix \ref{sec_experiment_details}. }\label{tab_compare_approximation_error}
{\small\begin{tabular}{@{}cccccccccc@{}}
\toprule
\textbf{Distance} & \textbf{Pythia-70M} & \textbf{BERT-Base} & \textbf{RoBERTa-Base} & \textbf{GPT-2} & \textbf{FLAN-T5-Base} & \textbf{BloomZ-560M}  \\ \midrule
0.05\% & $9_{\pm 1.4}\times10^{-4}$ & $3_{\pm 0.2}\times10^{-4}$ & $4_{\pm 0.4}\times10^{-4}$  & $2_{\pm 0.2}\times10^{-4}$ & $1_{\pm 0.2}\times10^{-4}$ & $2_{\pm 0.4}\times10^{-4}$\\
0.10\% & $1_{\pm 0.2}\times10^{-4}$ & $5_{\pm 0.9}\times10^{-4}$ & $5_{\pm 0.5}\times10^{-4}$  & $6_{\pm 0.7}\times10^{-4}$ & $8_{\pm 0.4}\times10^{-4}$ & $5_{\pm 1.5}\times10^{-4}$\\
0.15\% & $3_{\pm 0.3}\times10^{-4}$ & $9_{\pm 1.0}\times10^{-4}$ & $6_{\pm 0.9}\times10^{-4}$  & $8_{\pm 0.6}\times10^{-4}$ & $2_{\pm 0.3}\times10^{-4}$ & $9_{\pm 0.5}\times10^{-4}$\\
0.20\% & $4_{\pm 0.9}\times10^{-4}$ & $3_{\pm 0.4}\times10^{-3}$ & $9_{\pm 1.3}\times10^{-4}$  & $3_{\pm 0.3}\times10^{-3}$ & $3_{\pm 0.6}\times10^{-4}$ & $7_{\pm 0.6}\times10^{-4}$\\
0.25\% &$7_{\pm 1.4}\times10^{-3}$ &  $5_{\pm 1.4}\times10^{-3}$ & $5_{\pm 0.5}\times10^{-3}$  & $5_{\pm 0.4}\times10^{-3}$ &  $5_{\pm 1.2}\times10^{-3}$ & $5_{\pm 2.2}\times10^{-3}$\\ \midrule
\textbf{Distance} & \textbf{TinyLlama-1.1B} & \textbf{GPT-Neo-1.3B} & \textbf{OPT-1.3B} & \textbf{Gemma-2-2B} & \textbf{Mistral-7B} & \textbf{Llama-3-8B} \\ \midrule
0.05\% & $6_{\pm 0.5}\times10^{-5}$  & $3_{\pm 0.4}\times10^{-5}$ & $7_{\pm 0.2}\times10^{-5}$ & $4_{\pm 0.3}\times10^{-5}$ &  $9_{\pm 0.5}\times10^{-5}$ & $3_{\pm 0.3}\times10^{-5}$   \\
0.10\% & $1_{\pm 0.3}\times10^{-4}$ & $3_{\pm 0.3}\times10^{-4}$  & $7_{\pm 1.0}\times10^{-5}$ & $2_{\pm 0.1}\times10^{-4}$ &  $2_{\pm 0.2}\times10^{-4}$ & $6_{\pm 0.9}\times10^{-5}$  \\
0.15\% & $3_{\pm 0.7}\times10^{-4}$ & $6_{\pm 0.6}\times10^{-4}$  & $8_{\pm 0.2}\times10^{-5}$ & $3_{\pm 0.4}\times10^{-4}$ &  $3_{\pm 0.4}\times10^{-4}$ & $3_{\pm 0.6}\times10^{-4}$  \\
0.20\% & $4_{\pm 0.9}\times10^{-4}$ &  $1_{\pm 0.2}\times10^{-3}$ & $1_{\pm 0.1}\times10^{-4}$ &  $8_{\pm 0.7}\times10^{-4}$ &  $4_{\pm 0.2}\times10^{-4}$ & $4_{\pm 0.4}\times10^{-4}$   \\
0.25\% & $5_{\pm 0.8}\times10^{-3}$ & $5_{\pm 0.8}\times10^{-3}$ & $5_{\pm 0.1}\times10^{-3}$ &  $4_{\pm 0.5}\times10^{-3}$ & $4_{\pm 0.1}\times10^{-3}$  & $5_{\pm 0.4}\times10^{-4}$   \\
\bottomrule
\end{tabular}}
\end{table*}

Our algorithm involves two stages.
First, we run a meta-training procedure to obtain an initialization $\theta^{\star}$ by fine-tuning an LM on all the samples, similar to performing multitask learning \cite{wang2018glue}.
Second, we estimate model fine-tuning losses by solving a logistic regression problem to get $\hat \theta_{S_i}$ for every $i = 1, 2, \dots, m$.
Notably, the second estimation stage can be run entirely on CPUs, which will be very fast.

The key idea of why this works is a first-order approximation property that we have found empirically around the initialization LM.
The intuition is that for a highly over-parameterized network, the geometry around a local minimum solution tends to be flat \cite{zhang2024noise}, leading fine-tuning to behave like kernel regression locally \cite{malladi2023kernel}.
To aid this approximation, we hypothesize that after meta-training on all tasks, this initialization can adapt quickly to subsets of tasks.
This has also been observed in MAML \cite{finn2017model}, as depicted in Figure \ref{fig_pipeline}. 
The difference is that we further utilize the first-order approximation of large language models after meta-training. We have empirically observed that the linearization property holds across twelve LMs, as described next.

\subsection{Multitask training on all tasks}\label{sec_first_order_approximation}

Let the output of a network be $h_X(s, y)$, where $s$ is an input (e.g., sentence) and $y$ is the prediction label.
For example, in binary classification, $h_X$ is the log loss.
Here, $X\in \real^p$ denotes the trainable parameters of all layers, while $\theta^{\star}\in\real^p$ denotes trained parameters. We obtain a model initialization $\theta^{\star}$ by multitask training on all tasks. 

Note that $f(S)$ is equal to the average $h_{X}(s,y)$ over a set of samples $(s,y)$.
We examine first-order Taylor's expansion of $h_X(s, y)$ centered at $\theta^{\star}$:
\begin{align*}%
      h_X(s, y) \approx h_{\theta^{\star}}(s, y) &+ [\nabla_X h_{\theta^{\star}}(s, y)]^{\top} (X - \theta^{\star}) \\
      &+ \epsilon.
\end{align*}%

\emph{Our key observation is that $\epsilon$ remains negligible for $X$ around $\theta^{\star}$.}
We report empirical measurements of $\epsilon$ across twelve LMs, evaluated on the GLUE benchmark (with $n=9$ tasks) for BERT and RoBERTa, and the Alpaca dataset with $n=38$ tasks for the rest of ten LMs.
We use LoRA as the base fine-tuning procedure and obtain similar results to those with full fine-tuning.
We compute the relative residual sum of squares (RRSS):
\begin{align*}
\resizebox{.49\textwidth}{!}{
$\frac{\big(h_X(s,y) -  h_{\theta^{\star}}(s, y) - \nabla_X h_{\theta^{\star}}(s, y)^{\top} (X - \theta^{\star})\big)^2}{h_X(s,y)^2}.$
}
\end{align*}

We first verify that the fine-tuned models on subsets of tasks stay close to the initialization. We fine-tune one LLM on a random subset of tasks until the validation loss stops decreasing. We measure the relative distance from the fine-tuned model weights to the initialization, i.e., $\frac{\normFro{X-\theta^\star}}{\normFro{\theta^\star}}$ over 50 task subsets. We consider all model weights in the initialization as $\theta^\star$. 
For full model fine-tuning, the relative distance remains within 0.25\% on average. For LoRA fine-tuning, the relative distance stays within 0.19\%. 

Then, we report the first-order approximation error in Table \ref{tab_compare_approximation_error}. The results are averaged over $50$ randomly sampled subsets with fixed sizes (3 for GLUE and 19 for Alpaca).
Remarkably, across a range of relative distance values (between $X$ and $\theta^{\star}$) from 0.05\% to 0.25\%, the RRSS is at the range of $10^{-5}$ to $10^{-3}$.

\subsection{Efficient estimation for one task subset}

Next, we describe the inference of model fine-tuning performances on each subset $S_i$.
Importantly, we will achieve this using the functional values and the projected gradients obtained at the end of the first stage \emph{without} performing fine-tuning.

We illustrate the idea in binary classification, where $y$ is $1$ or $-1$. While the same works for multi-class and generative tasks. %
Consider the log-loss for binary classification:
\begin{align*}
    \ell(X) = \log\left( 1 + \exp\left(-y \cdot h_{X}(s, y) \right) \right).
\end{align*}%
Our key idea is to replace $h_X(s, y)$ above using $h_{\theta^{\star}}(s, y)$ plus the first-order term, and this is valid as long as $\epsilon$ is negligible, which is generally true for fine-tuning since
$X$ will be close to $\theta^{\star}$.
Let $b_s = - y \cdot h_{\theta^{\star}}(s, y)$ and let $g_s = \nabla h_{\theta^{\star}}(s, y)$.
We can approximate $\ell(X)$ with
\begin{align*}
    \hat\ell(X) = \log\left(1 + \exp\left(b_s - y\cdot g_s^{\top} (X - \theta^{\star}) \right) \right).
\end{align*}%

For a subset $S\subseteq\set{1, 2, \dots, n}$, let $\cD_S$ denote the combined samples from all of $S$, and let $n_S$ denote the total number of samples in $\cD_S$.
We estimate the model fine-tuning loss by minimizing the averaged $\hat\ell(X)$ over $X\in\real^p$:
\begin{align*}
    \frac{1}{n_S} \sum_{(s, y) \in \cD_S} \log\left(1 + \exp(b_s - y g_s^{\top} (X - \theta^{\star}))\right).
\end{align*}
Denote the above as $\hat L_S(X)$, which varies by $S$.
Let $\hat \theta_{S}\in\real^p$ be the solution from minimizing $\hat L_S(X)$.
In practice, the dimension of $X$ can be huge; thus, we apply random projection to reduce dimension, which can provably preserve the accuracy of the regression problem through the Johnson-Lindenstrauss lemma \cite{johnson1984extensions}.

We sample a $p$ by $d$ (e.g., for $d = 100$) Gaussian random matrix $P$ with each entry drawn from $N(0, d^{-1})$ and project the gradient as $P^{\top} h_{\theta^{\star}}(s, y)$.
We insert the projected gradient as $\tilde g$ into $\hat\ell(X)$.
Then, after solving the regression problem in dimension $d$, we map the minimizer $\hat X_d$ from dimension $d$ to $p$ using $P \hat X_d$.
This step only takes \emph{a few seconds}, which is extremely fast and is much faster than full fine-tuning, since it does not compute the gradient as both $b$ and $g$ are already computed after meta-training.
Thus, this step can be performed using CPUs.
Algorithm \ref{alg_estimate_performance} summarizes our procedure.

\begin{algorithm}[t!]
\raggedright
\caption{{\sc \textbf{GradEx}}: {\sc Fast \textbf{Es}timation} of LM Fine-Tuning Losses Using \textbf{Grad}ients}\label{alg_estimate_performance}
\textbf{Input}: $n$ training sets; 
 $m$ subsets $S_1, S_2, \dots, S_m$ of $\set{1, 2, \dots, n}$; a validation set of the target task\\
\textbf{Require:} LM $h_{\theta_0}$; Projection dimension $d$\\
{/*\qquad\qquad\quad~{\itshape \underline{Stage 1: Meta-training}} \hfill */}
\begin{algorithmic}[1]
    \State $\theta^{\star} \leftarrow$ fine-tune $h_{\theta_0}$ on $\cD_{\set{1,2,\dots,n}}$ 
    \State $P \leftarrow$ $p$ by $d$ isotropic Gaussian random matrix
    \For{$(s, y) \in \cD_{\set{1, 2, \dots, n}}$}
        \State $\tilde{g} \leftarrow P^{\top} \nabla h_{\theta^{\star}}(s, y)$ \Comment{project the gradient}%
        \State $b \leftarrow - y\cdot h_{\theta^{\star}}(s, y)$    
    \EndFor
\end{algorithmic}
{/*\qquad\qquad\quad~{\itshape \underline{Stage 2: Estimation}} \hfill */}
\begin{algorithmic}[1]
    \For{$i \leftarrow 1, \dots, m$}
        \State $\hat X_{d} \leftarrow$ $\min \hat L_{S_i}(X)$ with $\cD_{S_i}$ %
        \State $\hat \theta_{S_i} \leftarrow \theta^{\star} + P \hat X_d$
        \State $\hat f(S_i) \leftarrow$ evaluate {$h_{\hat \theta_{S_i}}$} on the target val set
    \EndFor
    \State Return ${\hat f(S_i), \text{ for every } i = 1, 2, \dots, m}$
\end{algorithmic}
\end{algorithm}

\subsection{Accelerating subset selection on LLMs}\label{sec_task_selection}

We now describe several use cases of the estimation algorithm, expanding on Examples \ref{ex_fs} and \ref{ex_re}.

\medskip
{\sc \textbf{GradEx-FS}}: During forward selection, we apply \algo{} to every subset encountered in the selection procedure.
After performing subset selection, we output the subset and use its data for augmentation.

\medskip
{\sc \textbf{GradEx-RE}}: %
For random ensembles, we first get a list of estimates $\hat f(S_1), \hat f(S_2), \ldots, \hat f(S_m)$ for every random subset. %
Then, compute a score $T_i$ for each task $i$ as the averaged result over all subsets that include $i$:
\begin{align}
    T_{i} = \frac{1}{n_{i}} \sum_{ 1\leq k \leq m:\ i \in S_k} \hat f\big(S_k\big), \text{ for } 1 \leq i \leq n, \label{eq_higher_order_affinity}
\end{align}
where $n_i$ is the number of subsets that include $i$.
Then, select a subset $\set{i \ | \ T_i < \gamma, i = 1, \ldots, n}$ using a threshold $\gamma$ adjusted via cross-validation.

\medskip
{\sc \textbf{GradEx-DS}}:
In data selection, one would like to pick a data subset from a collection of raw data. 
To apply our technique, one can preprocess the raw data by clustering it into $n$ groups. Then, use the above procedures to choose a subset from the $n$ groups of data samples.

\paragraph{Examples.}
We illustrate our approach in a noisy addition example.
Consider adding two digits of length $5$:
\begin{flushleft}
    \textbf{IN}~~~\quad\quad\quad\quad \emph{\underline{6 7 0 1 3}}~~ + ~~\emph{\underline{2 3 9 2 4}}\\
    \textbf{OUT}~~ \emph{0 \underline{7}} \,|\, \emph{0 \underline{3 7}} \,|\, \emph{0 \underline{9 3 7}} \,|\,\,\emph{1 \underline{0 9 3 7}} \,|\, \emph{\underline{9 0 9 3 7}}
\end{flushleft}
One can write down the intermediate calculations plus carry bits.
We thus generate a synthetic addition set by generating input pairs of length $5$ between $0$ and $9$.
We create $20$ groups; ten are correct, while the rest involve randomly generated output digits.
Then, we aim to separate the noisy groups from the correctly-labeled groups.

\begin{figure}[h!]
    \centering
    \includegraphics[width=0.4805\textwidth]{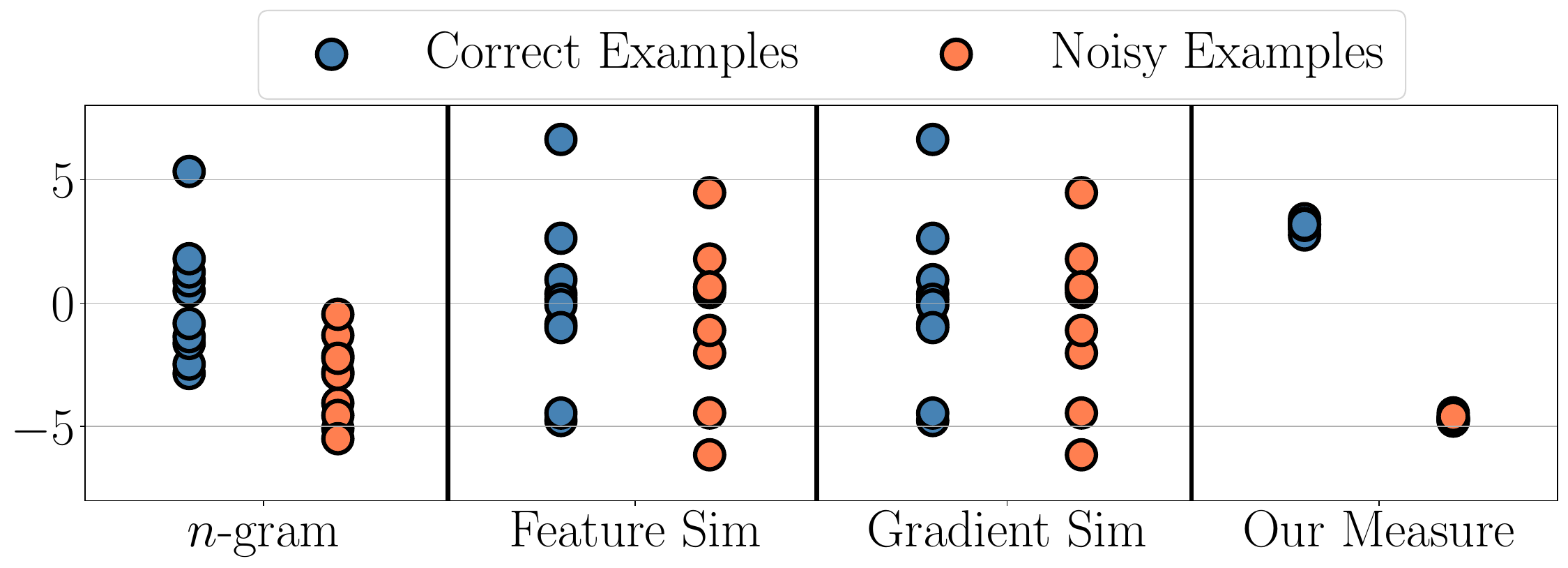}
    \caption{Illustrating the separation between correctly-labeled and noisy examples, using estimated values of $T$ in equation \eqref{eq_higher_order_affinity}, compared to several existing measures.
    Our measure from $T$ (rightmost) leads to a much clearer separation of correct vs. noisy examples compared to the three measures.
    In particular, the second and third figures utilize gradient-based feature similarity, and our loss-based selection procedure yields a more accurate separation than these measures.}\label{fig_relevance_scores}
\end{figure}

We apply \algore~to train a GPT-2 transformer network. %
For comparison, we also report the results from measuring $n$-gram features \cite{xie2023data}, feature similarities inside the transformer \cite{ivison2023data}, and gradient similarities of the fine-tuned model \cite{xia2024less}.
We report our findings in Figure \ref{fig_relevance_scores}, showing that our procedure can lead to a better separation than the baseline measurements. %

\subsection{Comparison with prior methods}

We discuss the memory and runtime complexity of our algorithm. 
Regarding memory requirement, our algorithm matches the base fine-tuning method and inherits the same number of trainable parameters as LoRA.
The runtime of our algorithm involves meta-training on $n$ tasks, computing the gradients of all the $n$ tasks, and solving logistic regression in dimension $d$ on $m$ subsets.

In particular, our algorithm reduces $m$ model fine-tuning runs to a single meta-model training.
As mentioned earlier, the additional estimation stage incurs very little overhead; after dimension reduction, solving each logistic regression problem takes less than 2 seconds per task subset. This estimation stage takes less than 10\% of the total computation cost in our overall procedure.
Regarding the meta-training stage, our algorithm incurs comparable computational costs to existing data selection methods, such as those presented by \citet{xie2023data} and \citet{xia2024less}.
Table \ref{tab_related_works_comparison} summarizes this comparison, including the number of forward passes required by each approach. 
Note that the number of forward passes is for one training step. For every method, the number of backward passes equals the forward passes. For the calculation of each method, see Appendix \ref{sec_algorithm_details} for the details.

\begin{table}[t!]
\centering
\caption{Summary of runtime complexity between our algorithm and existing solutions for subset selection, as a function of the number of data sources $n$. We describe the constants in terms of the forward passes each method takes. Here, $\alpha$ denotes the size of subsets sampled in random ensembles.}\label{tab_related_works_comparison}
\resizebox{\columnwidth}{!}
{\begin{small}
\begin{tabular}{@{}l c c c c @{}}
\toprule
\textbf{Methods} &  \makecell{\textbf{Runtime}} & \textbf{\# Forward} \\ \midrule %
Forward Selection (FS)  & $O(n^3)$ & $\frac{1}{6} n^3$ \\ 
Random Ensemble (RE)   & $O(n \log n )$ & $\alpha n \log n $\\ %
DSIR \cite{xie2023data}  & $O(n)$   & $n$  \\ %
DEFT \cite{ivison2023data}  & $O(n)$ & $3n$ \\ %
LESS \cite{xia2024less}    & $O(n)$ & $3n$  \\ %
\algofs  & $O(n)$ & $3n$  \\ %
\algore  & $O(n)$ & $3n$  \\ %
\bottomrule
\end{tabular}
\end{small}}
\end{table}

%% file: experiment.tex
\begin{table*}[t!]
\centering
\caption{We report the relative RSS between $\hat f(S)$ and $f(S)$, measured on Alpaca and StrategyQA. For measuring speedup, we report the ratio of the number of FLOPs required for computing $f(S)$ to the number of FLOPs required by \algo{}. The speedup remains the same across different models for the same dataset, as it stems from a reduced number of forward and backward passes.}\label{table_compare_distance}
{\small\begin{tabular}{@{}cccccc|c@{}}
\toprule
Alpaca & {\textbf{GPT-2}} & {\textbf{FLAN-T5-Base}} & {\textbf{TinyLlama-1.1B}} & {\textbf{GPT-Neo-1.3B}} & {\textbf{Llama-3-8B}} & \textbf{Speedup} \\ \midrule
\algofs{} & $7.4 \times 10^{-4}$  & $3.2 \times 10^{-4}$  & $3.9 \times 10^{-4}$  & $3.5 \times 10^{-4}$  & $2.7 \times 10^{-4}$  & 17.6$\times$\\
\algore{} & $8.7 \times 10^{-4}$& $3.4 \times 10^{-4}$& $4.2 \times 10^{-4}$& $3.8 \times 10^{-4}$& $2.9 \times 10^{-4}$ & 43.3$\times$ \\ 
\midrule
StrategyQA & {\textbf{GPT-2}} & {\textbf{FLAN-T5-Base}} & {\textbf{TinyLlama-1.1B}} & {\textbf{GPT-Neo-1.3B}} & {\textbf{Llama-3-8B}} & \textbf{Speedup} \\ \midrule
\algofs{} & $7.4 \times 10^{-3}$  & $3.0 \times 10^{-4}$   & $3.2 \times 10^{-4}$   & $3.0 \times 10^{-4}$   & $2.4 \times 10^{-4}$ & 30.5$\times$\\
\algore{} & $8.9 \times 10^{-3}$  & $3.6 \times 10^{-4}$   & $3.8 \times 10^{-4}$   & $3.4 \times 10^{-4}$   & $2.8 \times 10^{-4}$ & 44.8$\times$\\
\bottomrule
\end{tabular}}
\end{table*}

\section{Experiments}\label{sec_exp}

We now validate \algo{} and its use cases across various datasets and models, focusing on the following key questions.
Does the estimation procedure accurately approximate the true model fine-tuning losses?
How much computational cost does it save relative to classical subset selection methods that require repeated model fine-tuning?
How effective are the selection methods based on estimated results in downstream evaluation?

Our experiments show that \algo{} approximates fine-tuned model losses within \textbf{1}\% error, tested on five different LMs, including Llama-3-8B. \algo{} reduces the number of FLOPs by up to \textbf{43}$\times$ and GPU hours by \textbf{46}$\times$ compared to conventional subset selection. 
Next, we evaluate \algo{} for \textit{instruction tuning} and \textit{chain-of-thought fine-tuning}. 
In both cases, our algorithm performs on par with conventional subset selection that uses true fine-tuning results while incurring less than \textbf{0.5}\% computation costs.
Our algorithm outperforms existing selection methods \cite{xie2023data,xia2024less} on ToxiGen by \textbf{3.8}\% and TruthfulQA by \textbf{2.4}\%, while using comparable computation costs.%

\subsection{Experimental setup}

Our algorithm is broadly applicable to estimating the performance of LLM fine-tuning. In this section, we focus on instruction tuning and chain-of-thought fine-tuning.
For the former, given a set of source tasks and a target task, we aim to select source tasks relevant to the target task.
For the latter, we fine-tune an LM to generate chain-of-thought reasoning steps and the answer to a question. Several explanations are possible, while some of them are incorrect. We aim to select explanations pertinent to the reasoning task using subset selection.

For \textit{instruction tuning}, we use three datasets, including FLAN V2 with 1,691 tasks, Chain of Thoughts with 18 tasks, and Alpaca with 38 tasks. These datasets encompass over 150 task categories. See Appendix \ref{sec_experiment_details} for the statistics.

For \textit{chain-of-thought fine-tuning}, we use CommonSenseQA \cite{talmor2019commonsenseqa} and StrategyQA \cite{geva2021did} datasets. The chain-of-thought explanations are generated with a GPT-3 175B model. We sample 10\% of the data for evaluating $f(S)$. We partition the data into $100$ groups by applying spectral clustering to a matrix of gradient cosine similarity scores between every pair of samples. Each group refers to one task. In the end, we have $100$ tasks from the $100$ groups in both the CommonSenseQA and StrategyQA datasets.

We set LoRA as the base protocol. We adjust the rank parameter of LoRA between 16, 32, 64, and 128. For random ensembles, we sample $1000$ subsets, each containing 75\% of all tasks, to ensure the $T$ scores (cf. equation \eqref{eq_higher_order_affinity}) have converged.

\subsection{Results on approximation errors}\label{sec_running_example}

We now assess the accuracy of \algo{}. 
We measure the relative error between the estimated $\hat f(S)$ and the true value of $f(S)$ (which can be obtained by running full fine-tuning) as the averaged error over $m$ subsets:
\[ \frac{1}{m} \sum_{i=1}^m \frac{( f(S_i) - \hat f(S_i) )^2}{( f(S_i))^2}.\]
More specifically, we obtain $f(S)$ by fine-tuning a pretrained LM on the samples of $S$ along with the samples from the target task.
We measure computation cost in terms of the total number of floating-point operations (FLOPs) and the number of GPU hours (measured on a desktop with three Nvidia RTX 6000 GPUs).

We evaluate the relative distance on both Alpaca and StrategyQA using five different LMs listed in Table \ref{table_compare_distance}.
Due to computation constraints, for models with more than 1 billion parameters, we sample $100$ subsets to estimate the approximation error for the random ensemble.

\begin{figure*}[t!]
    \centering
    \begin{subfigure}[b]{0.48\textwidth}
    \begin{minipage}[b]{0.49\textwidth}
        \centering
        \includegraphics[width=\textwidth]{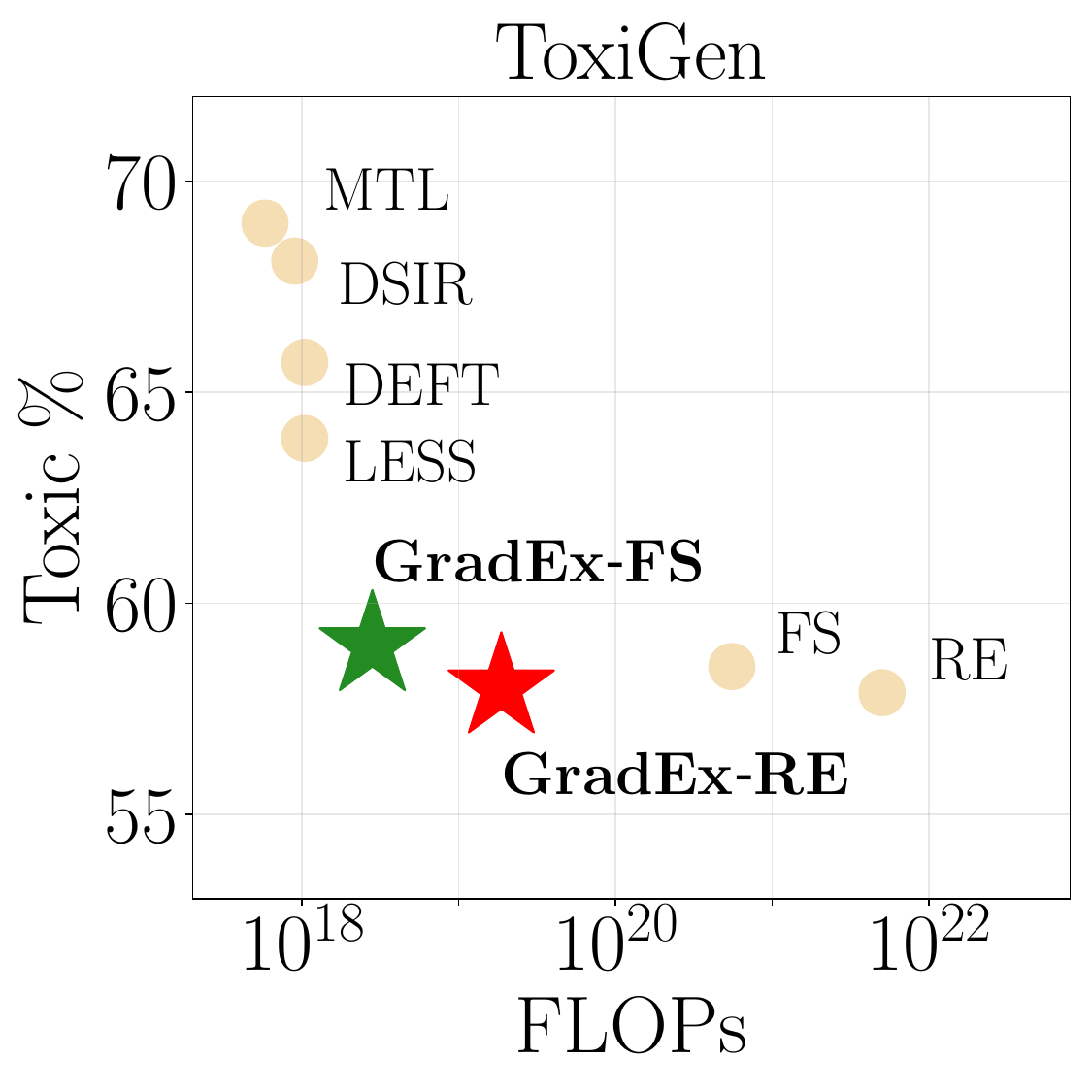}
    \end{minipage}
    \begin{minipage}[b]{0.49\textwidth}
        \centering
        \includegraphics[width=\textwidth]{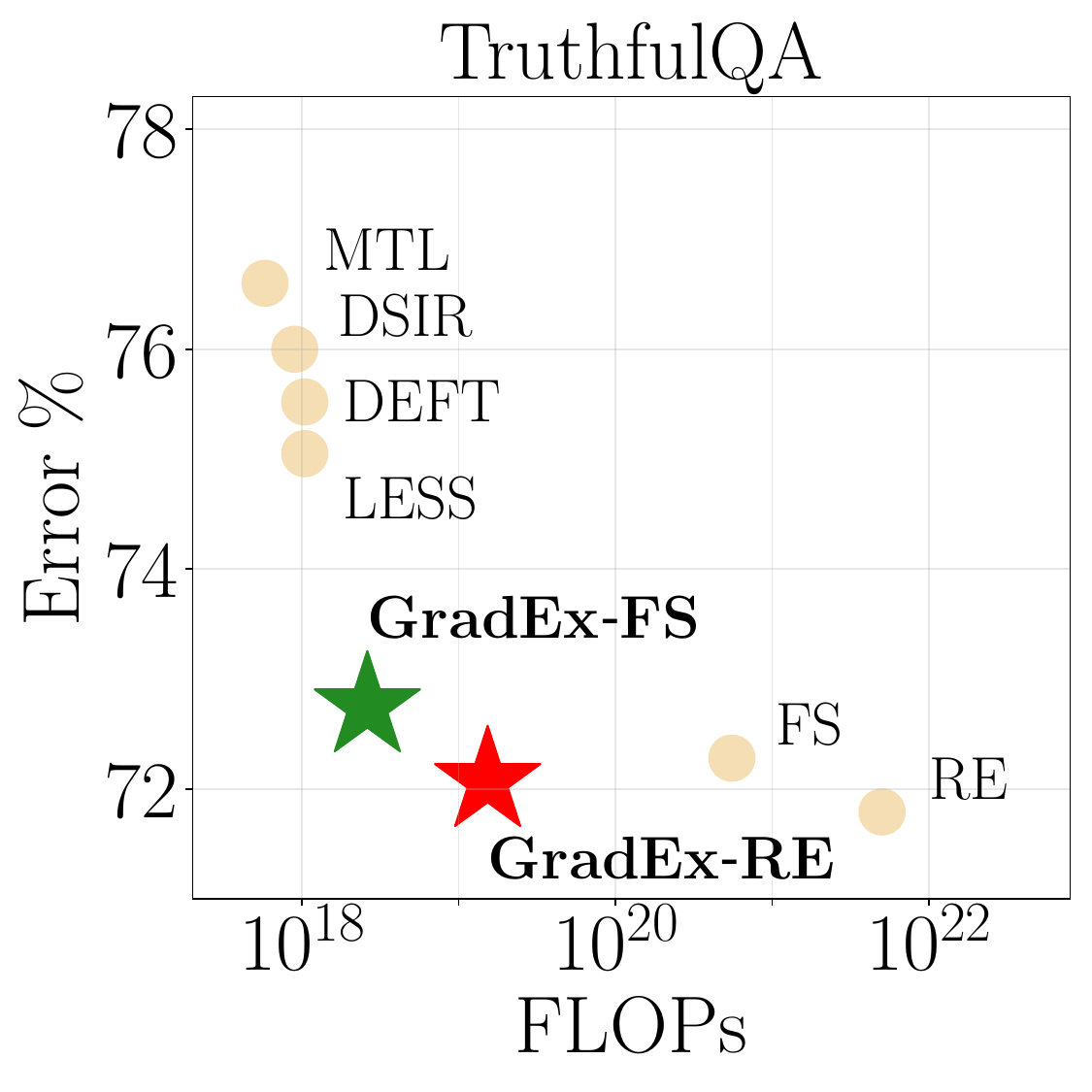}
    \end{minipage}
    \subcaption{Task selection for instruction tuning ($n=1,729$)}\label{fig_compare_task_selection}
    \end{subfigure}
    \begin{subfigure}[b]{0.48\textwidth}
    \centering
    \begin{minipage}[b]{0.49\textwidth}
        \centering
        \includegraphics[width=\textwidth]{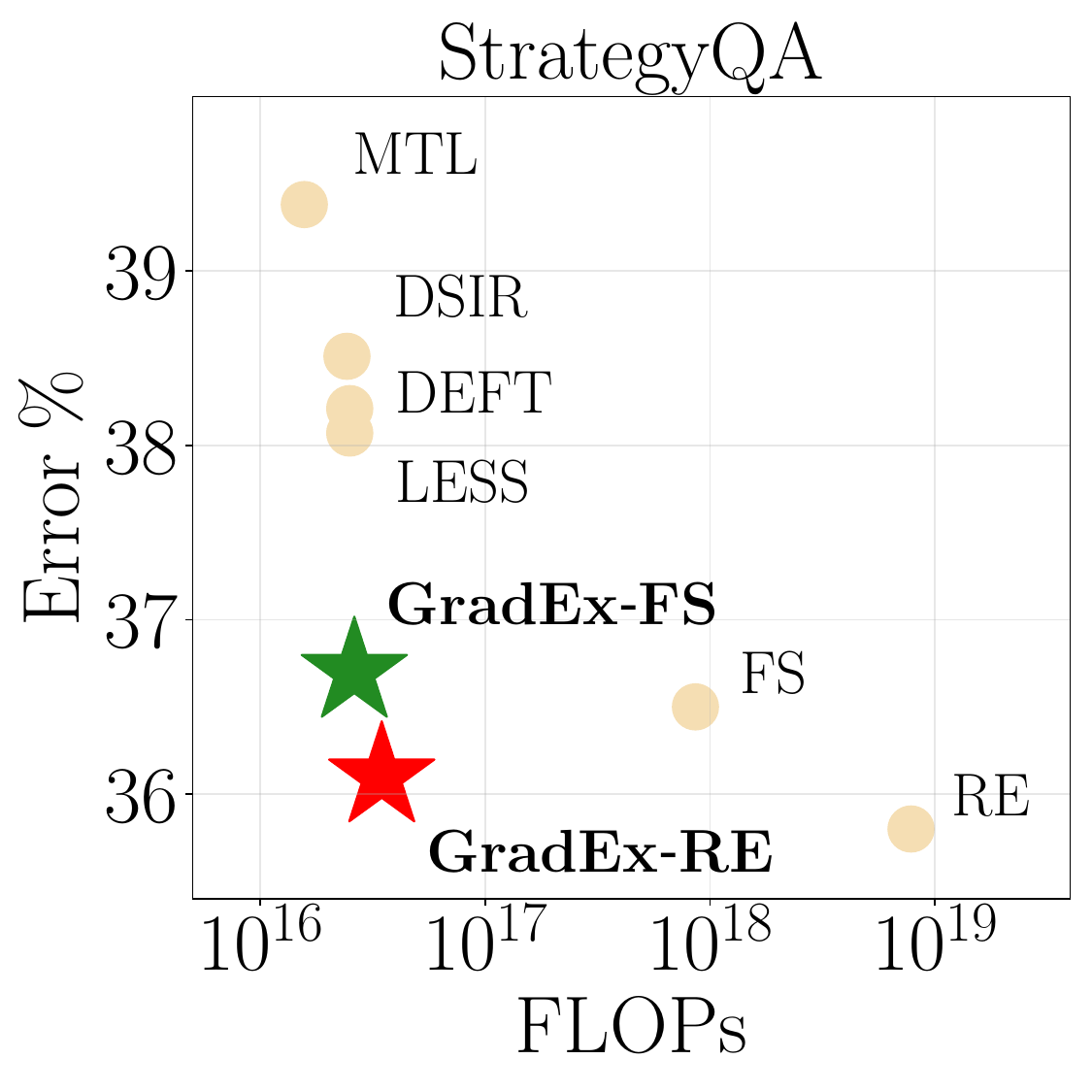}
    \end{minipage}
    \begin{minipage}[b]{0.49\textwidth}
        \centering
        \includegraphics[width=\textwidth]{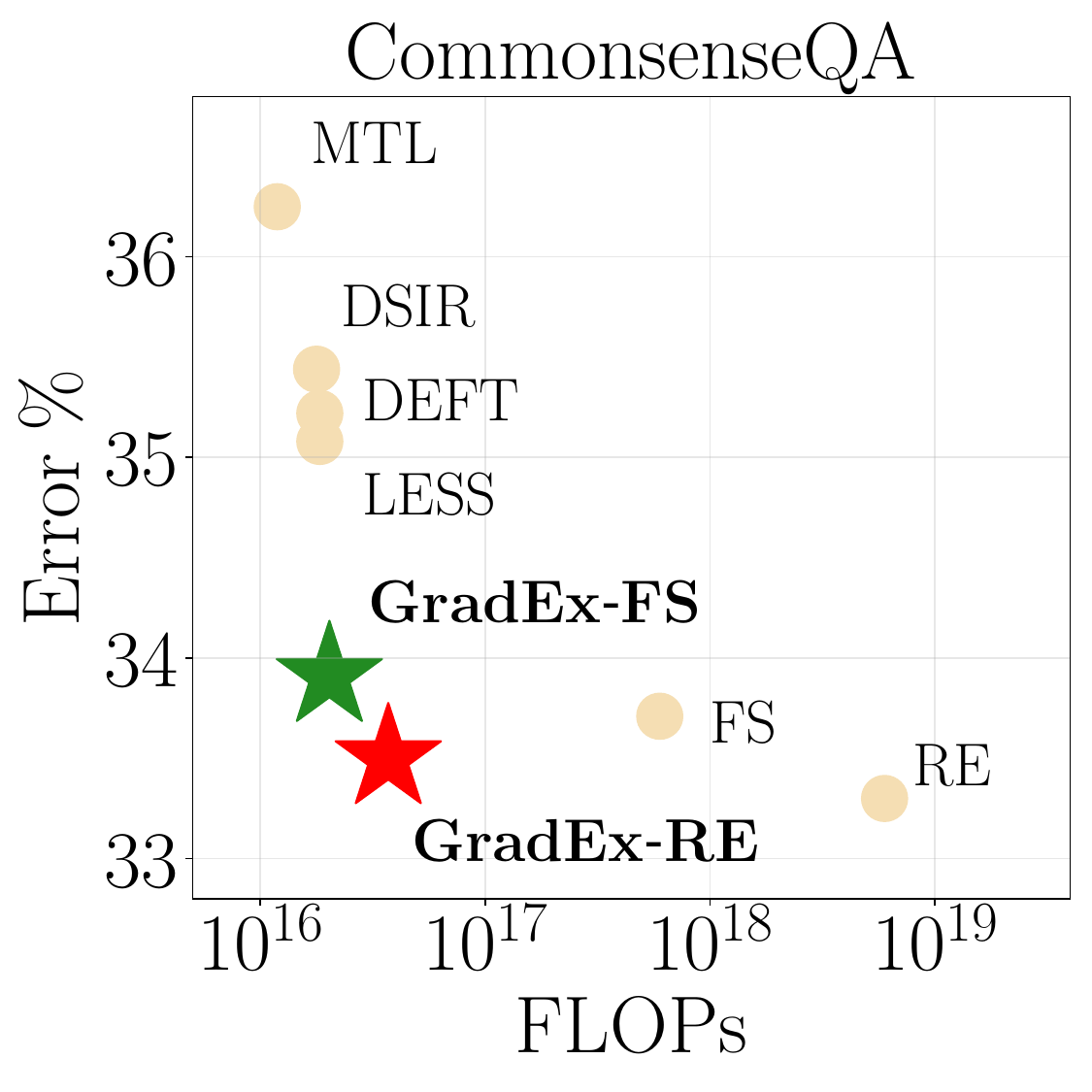}
    \end{minipage}
    \subcaption{Data selection for chain-of-thought fine-tuning ($n=100$)}\label{fig_compare_runtime_performance_reasoning}
    \end{subfigure}
    \caption{Illustration of the tradeoff between the number of FLOPs (computation cost) and test error rate, measured on our methods and six baseline methods. \emph{Our approach delivers comparable downstream performance to conventional subset selection methods}, while using less than \textbf{0.5}\% of total computation. 
    Furthermore, our approach achieves a \textbf{3.8}\% performance improvement over existing data selection methods using comparable computation costs.
    In particular, MTL refers to the standard approach of training a single model on the data from all tasks, and we can see that our approach indeed reduces the error rates compared to MTL.
    The comparison of GPU hours is qualitatively similar and can be found in Appendix \ref{sec_additional_exp_results}.}\label{fig_ill_tradeoff}
\end{figure*}

We find that our algorithm approximates the fine-tuned model losses within \textbf{1}\% error across the five LMs. Furthermore, the approximation quality is generally better for larger models.

On Alpaca, we find that \algofs{} uses 117 GPU hours for Llama-3-8B; This is \textbf{17.6}$\times$ less computation compared to running forward selection with true fine-tuning. As for random ensembles, \algore{} uses 120 GPU hours for Llama-3-8B and \textbf{43.3}$\times$ less computation compared to true fine-tuning.
We also observe qualitatively similar results when measured on StrategyQA: \algo{} achieves \textbf{30.5}$\times$ and  \textbf{44.8}$\times$ less computation compared to true fine-tuning in the forward selection and random ensembles, respectively.

\begin{figure}[t!]
    \begin{subfigure}[b]{0.240\textwidth}
        \centering
        \includegraphics[width=0.975\textwidth]{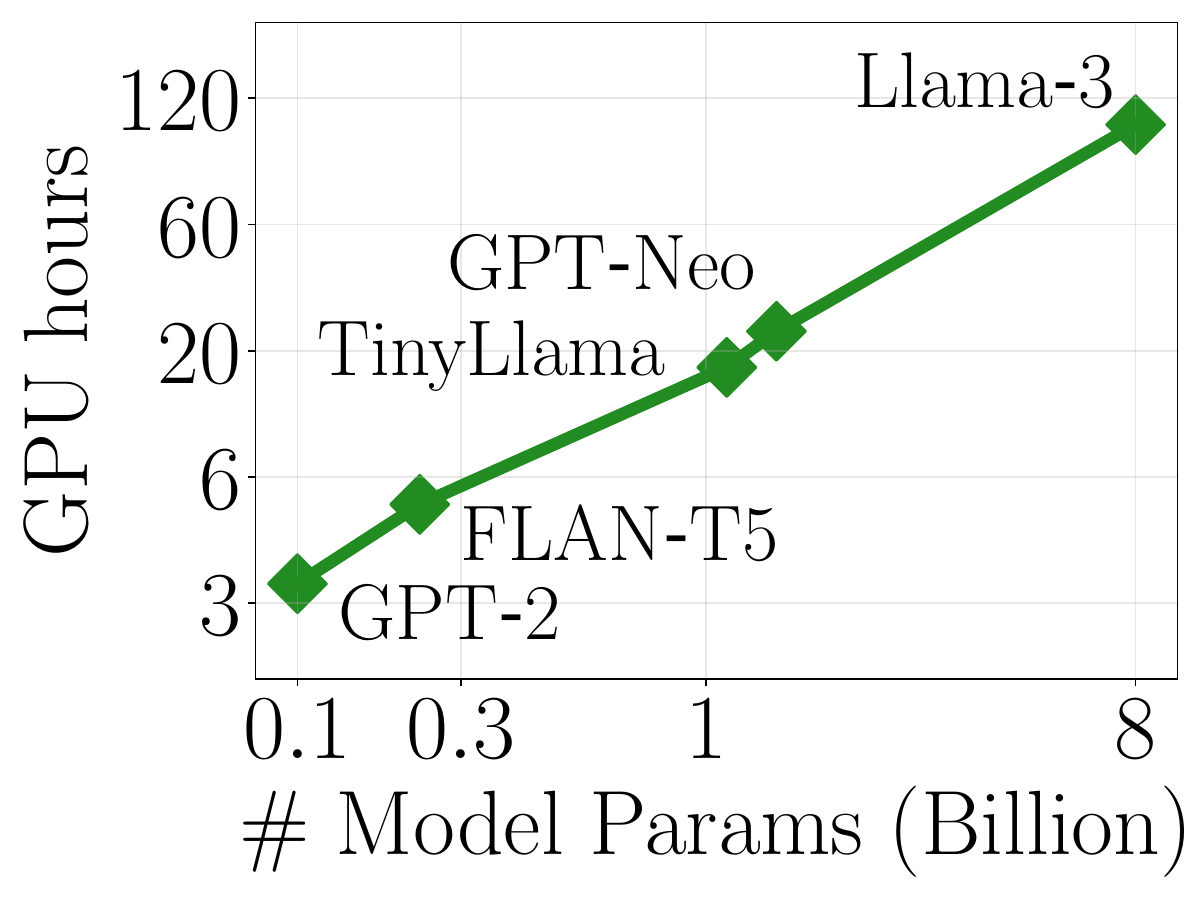}
        \subcaption{Our estimation (\algo{})}\label{fig_rt_ge}
    \end{subfigure}%
    \begin{subfigure}[b]{0.240\textwidth}
        \centering
        \includegraphics[width=0.975\textwidth]{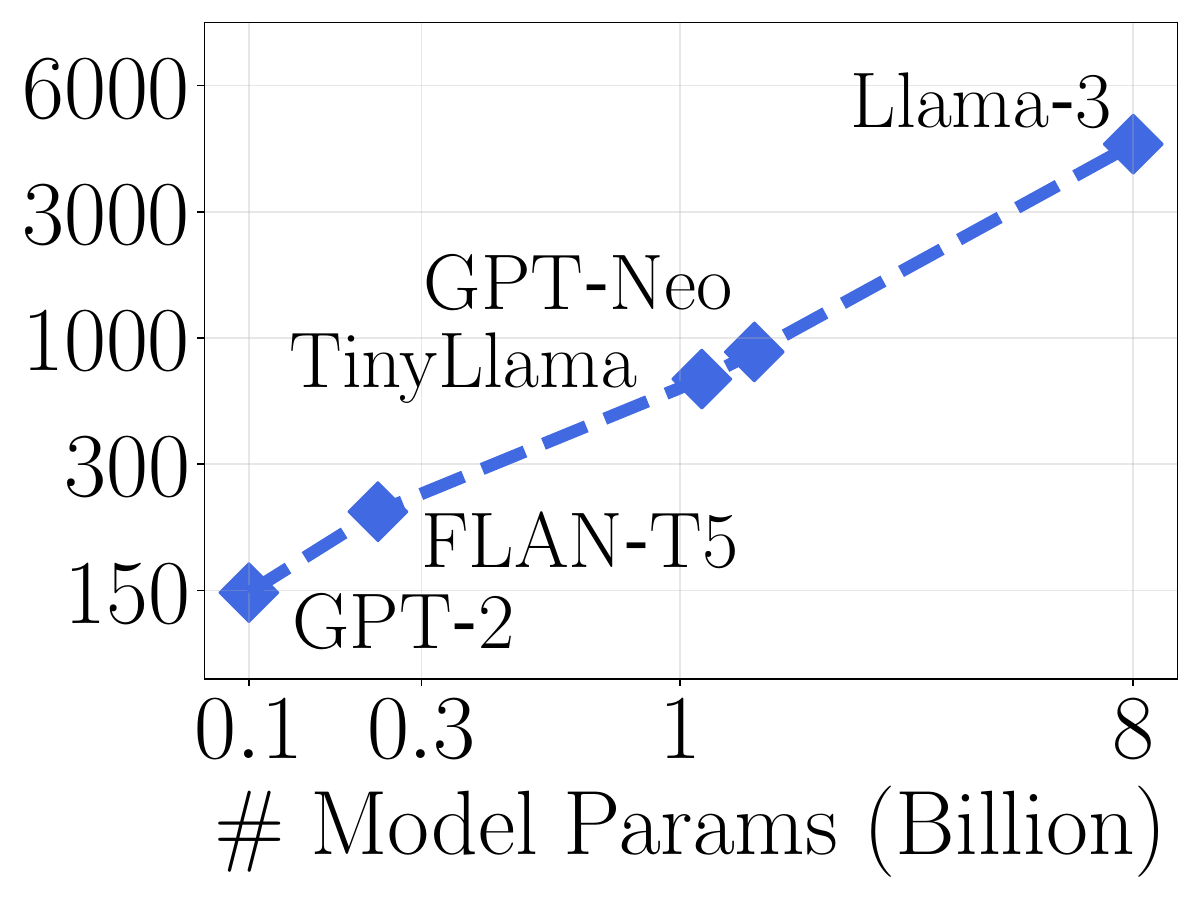}
        \subcaption{Full fine-tuning}\label{fig_rt_ft}
    \end{subfigure}
    \caption{Number of GPU hours as the number of model parameters between our estimation approach (left, Figure \ref{fig_rt_ge}) and full fine-tuning (right, Figure \ref{fig_rt_ft}). We estimate the full fine-tuning cost by fine-tuning on randomly sampled 100 subsets of tasks.}\label{fig_scaling}
\end{figure}

We note that the speedup remains consistent across different models when applied to the same dataset. Figure \ref{fig_scaling} illustrates the number of GPU hours used by \algo{} when running our approach on five LMs on Alpaca vs. full fine-tuning.

\medskip
\noindent\textbf{Projection dimension:} Recall that we project the dimension of the gradients down to a much smaller dimension during the second estimation stage. We vary the projection dimension $d$ between 50, 100, 200, and 400 for testing \algo{} on FLAN-T5-Base. We observe that once $d$ increases above $100$, the error stabilizes around {0.03\%}. Hence, we set $d$ as $100$ for all the experiments. With $d=100$, solving a logistic regression problem for one subset takes less than 2 seconds.

\medskip
\noindent\textbf{Reducing overfitting in meta-training:} Recall that in the meta-training stage, we apply multitask training to the combined samples of all the tasks. We conduct a preliminary experiment where we use a sharpness-reduced training algorithm (i.e., sharpness-aware minimization \cite{foret2020sharpness,zhang2024noise}) for multitask training (in place of SGD). We find that this can reduce the approximation error (relative to full fine-tuning results) by 23\%. Further applying multitask and meta-learning techniques to improve \algo{} would be an interesting question for future work.

\subsection{Results on model fine-tuning} \label{sec_finetuning_evaluation}

We compare our algorithms to forward selection (FS) and random ensemble (RE) with full fine-tuning.
Additionally, we compare with four baseline methods, including fine-tuning a single model on all tasks with LoRA (MTL), Data Selection with Importance Resampling (DSIR) \cite{xie2023data},
Data-Efficient Fine-Tuning with cross-task nearest neighbors clustering based on feature similarity (DEFT) \cite{ivison2023data},
and Low-rank gradiEnt Similarity Search (LESS) \cite{xia2024less}.
For each baseline, performance is reported based on fine-tuning a pre-trained model with the same number of trainable parameters on the selected subset of tasks.

For instruction tuning, we follow the protocol of \citet{wang2023far} to evaluate the toxicity and truthfulness of fine-tuned models in ToxiGen \cite{hartvigsen2022toxigen} and TruthfulQA \cite{lin2021truthfulqa}, respectively. For ToxiGen, we measure the percentage of toxic outputs generated by the model. For TruthfulQA, we evaluate model accuracy for identifying truthful statements. For StrategyQA and CommonsenseQA, we measure the accuracy of the generated answers.

\paragraph{Results for instruction tuning.}%
We illustrate the results of applying \algofs{} and \algore{} to select tasks in instruction tuning in Figure \ref{fig_compare_task_selection}, using the TinyLlama-1.1B model. 
For ToxiGen, we plot the percentage of toxic generations. For TruthfulQA, we plot the error rate as one minus the accuracy of identifying the correct answer.
For each method, we vary the ratio of selected tasks between 5\%, 10\%, 15\%, and 20\%, and report the best result.

First, we compare \algofs{} with directly testing the pretrained LM and fine-tuning the last output layer of the LM (Feature Transfer). We find that \algofs{} outperforms both methods by 9\% and 11\% on average, respectively.
Compared to fine-tuning a model on all tasks (MTL), \algofs{} leads to an 8\% improvement. 

Second, \algofs{} and \algore{} deliver comparable performance (within a 1\% performance gap) to the forward selection and random ensemble, using less than \textbf{0.5}\% of the computation cost of full fine-tuning. 

Lastly, we find that \algofs{} improves over DSIR, DEFT, and LESS by \textbf{3.8}\%, all of which use a similar number of FLOPs, and \algore{} outperforms these baseline methods by \textbf{4.7}\%.

\paragraph{Results for chain-of-thought fine-tuning.} %
Next, we report the results on chain-of-thought fine-tuning, illustrated in Figure \ref{fig_compare_runtime_performance_reasoning}, tested on FLAN-T5-Base.
In summary, our findings are consistent with those of instruction tuning.

First, \algofs{} outperforms direct testing of the pre-trained model and feature transfer by 7\% and 16\%, respectively.
\algofs{} also improves the accuracy by 5\% over MTL.
Second, even with estimation, we observe similar downstream performance to forward selection and random ensemble (with true fine-tuning) while using only \textbf{3}\% and \textbf{0.5}\% of the total computational costs.
Lastly, both \algofs{} and \algore{} can outperform the four baseline selection methods by \textbf{2.4}\% and \textbf{3.5}\% (on average), respectively.

As for data selection, we first cluster the samples into $n$ groups. %
We vary $n$ between $50$, $100$, $200$, and $400$. We also evaluate our approach without the clustering step.
We find that using $100$ groups yields the best performance and indeed outperforms not using clustering by 1.4\%.
Therefore, we set the number of groups as $100$ in the chain-of-thought fine-tuning.

%% file: related.tex
\section{Related Work}

\textbf{Parameter-efficient fine-tuning:}
One influential line of work has sought to design adapters, which are small modules injected into the intermediate layers of a deep neural network.
With adapters, only a small fraction of the entire network has to change inside the adapters, and this approach has found applications in many settings such as text classification \cite{howard2018universal}, text transfer \cite{pfeiffer2021adapterfusion}, and cross-lingual transfer such as named entity recognition and commonsense reasoning \cite{bapna2019simple}.
Another different approach is to use LoRA \cite{hu2021lora}, which constrains the fine-tuning region inside a low-rank subspace, greatly improving the parameter efficiency of fine-tuning.
Both LoRA and follow-up works, such as QLoRA \cite{dettmers2024qlora} and ReLoRA \cite{lialin2023relora}, focus on fine-tuning a single task.
\citet{mahabadi2021parameter} studies parameter-efficient multitask fine-tuning via shared hypernetworks.
Our work can be viewed as expanding these methods to multitask learning (MTL). We believe this connection between MTL and fine-tuning would be worth further exploration in future work.

\medskip
\noindent\textbf{Multitask learning for NLP:}
There is also a line of work on building transfer learning approaches for tackling low-resource languages and tasks.
For instance, adapting from a high-resource language to another low-resource language is a particularly effective strategy \cite{neubig2018rapid}.
\citet{vu2020exploring} explore the transferability using a large collection of NLP tasks. This large-scale analysis underscores the intricacy of representational transfer in natural languages.
\citet{vu2021spot} examine feature transfer in the context of prompt tuning.
There is another line of work for domain adaptation using a mixture of experts \cite{shazeer2016outrageously}.
\citet{wu2019understanding} and \citet{yang2020precise} provide a theoretical analysis of the multi-headed architecture commonly used for conducting multitask learning.
Their work highlights the issue of negative transfers in multitask learning.
For further references, see recent surveys discussing the ongoing developments and challenges of multitask learning for NLP \cite{raffel2020exploring,zhang2023survey}.

\medskip
\noindent\textbf{Data modeling:}
There is a growing line of work on understanding the role of individual samples in deep networks and large models.
\citet{kozareva2011class} propose a centrality and regularization-based algorithm by leveraging graph structures to explore the relationship between multiple instances in a semantic network.
\citet{wettig2024qurating} select pre-training data by training a rater model to evaluate four data quality criteria that align with human intuitions.
In contrast, this work focuses on task-relatedness in fine-tuning language models. This is a simpler problem than pre-training, as the number of tasks is smaller. A potential adaptation could be splitting pre-training into multiple phases and applying our techniques to develop a curriculum for pre-training.

Our approach is also related to the work of \citet{ilyas2022datamodels} and \citet{park2023trak} on using random sampling for data modeling and attribution.
The difference is that we utilize the linearization property of LLMs for fine-tuning close to the initialization.
A closely related work by \citet{li2023identification} introduces a surrogate modeling approach to identify subsets of source tasks that benefit a target task. They show that a linear surrogate model is particularly helpful for identifying negative transfers, including higher-order negative transfers from a subset of source tasks to the target task.
This higher-order transfer corresponds to a boosting procedure that turns out to be very helpful in practice \cite{li2023boosting}.
In addition, this surrogate modeling approach can also be used in data augmentation for finding tree-structured compositions \citet{lilearning}.
We remark that in a follow-up paper, we provide an improved version of our algorithm for low-rank adapter tuning \cite{li2025efficient}.
Moreover, we find that for many LLMs, one does not need the meta-training stage to apply such low-rank adaption methods before using the linearization technique.
It would be very interesting to understand the underlying mathematical structure of LLMs that enables such geometric properties in practice \cite{zhang2023mathematical}.

\section{Conclusion}

This paper introduced a novel method for estimating LM fine-tuning performances with a first-order approximation approach.
The method can significantly accelerate conventional subset selection, thus unlocking its applications to task/data selection for fine-tuning. Evaluation across numerous datasets and LMs demonstrates the benefit of \algo{} when applied to subset selection.

\section*{Acknowledgments}
Thanks to the anonymous referees and the action editor for their constructive feedback.
The work of D. Li, Z. Zhang, and H. Zhang is partly supported by NSF award IIS-2412008.
D. Li was also partially funded by a PhD fellowship from JPMorgan Chase \& Co. Any views or opinions expressed herein are solely those of the authors listed, and may differ from the views and opinions expressed by JPMorgan Chase \& Co. or its Affiliates. %

\section*{Limitations}

One limitation of our algorithm is that it requires estimating the fine-tuned model's performance based on all the model weights. Designing techniques to estimate fine-tuning performance in a limited set of model weights or for closed-source models such as GPT-4 is an interesting open question. 
Our study is also restricted to fine-tuning, while there may well be other scenarios where one would like to adopt a language model, such as in-context learning or alignment with RLHF. Further exploring functional approximation techniques for these settings would be another promising avenue for future work.

\section*{Broader Implications}

This paper examines the problem of fine-tuning language models given multiple data sources. While the use of language models may have potential societal consequences in the future, there are no specific ethical concerns arising from our work. Due to the technical nature of this paper, there are no direct implications of negative societal impacts.

%% file: appendix.tex
\section{Omitted Materials from Section \ref{sec_motivation}}

\textbf{Notations.} %
We provide a list of notations and their meanings used in the paper for reference below:
\begin{itemize}
    \item $S$: A subset of the $n$ tasks from $\set{1, 2, \dots, n}$.
    \item $f(S)$: {The performance of an LM fine-tuned on tasks in $S$, evaluated on the  target set}.
    \item $\theta^\star$: {The meta-initialization, i.e., vectorized LM weights fine-tuned on all tasks}.
    \item $\hat\theta_S$: {The vectorized LM weights fine-tuned on a subset of tasks $S$}.
    \item $h_{\theta^\star}(s, y)$: Model loss given an input pair $s, y$.
    \item $\nabla h_{\theta^\star}(s, y)$: {Vectorized gradients of model output with respect to model weights $X$}.
    \item $T_{i}$: {The average performance of $f(S)$ over multiple subsets $S$ that include task $i$}, for every $i = 1, 2, \dots, n$.
\end{itemize}

\paragraph{$f(S)$ is not monotone.} We find that $f(S)$ is not monotone, in the sense that if we add one helper task into $S$, this does not necessarily improve the outcome. Concretely, we start with an initial set $S$. It contains a target task $t$ corresponding to the instructions of ``arrange'' in Alpaca. 
Then, we keep adding a ``helping'' task ($i$) to $S$, if $f(\set{i, t}) < f(\set{t})$ (reducing the loss).
According to our experiment, once more than two tasks are added, $f(S)$ increases gradually, indicating that adding more tasks can worsen the performance of the target task, and these are all ``helpful'' tasks (in the sense of pairwise transfer).

\paragraph{$f(S)$ is not submodular.} We also find that $f(S)$ is not submodular. A function $f(\cdot)$ is submodular if for any two subsets $S \subseteq S' \subseteq\set{1, 2, \dots, T}$ and any single task $x$, $f(\{x\} \cup S') - f(S') \le f(\{ x \} \cup S) - f(S)$.
To test this, we start with a set $S$ that includes the target task and a task that negatively transfers, i.e., increases the loss of the target task. Then, we also keep adding a ``helping'' task ($i$) to $S$.
According to our experiment, we observe that the benefit of adding tasks gradually decreases. Initially, adding the first two tasks to the original pair improves performance. However, this improvement diminishes once three or more tasks are added.

\section{Omitted Materials from Section \ref{sec_method}}\label{sec_algorithm_details}

\textbf{Meta-training improves approximation quality.} In Section \ref{sec_method}, we observe that the first-order expansion near a meta-initialization fine-tuned on all tasks provides an accurate approximation. Next, we evaluate whether this approximation can be achieved without the meta-training step. We measure the first-order Taylor expansion on the pretrained GPT-Neo-1.3B and TinyLlama-1.1B models. We observe that the approximation from the pretrained initialization incurs within 7\% error. In contrast, using meta-initialization is significantly better. The approximation achieves less than 1\% error.  We report the results in Table \ref{tab_compare_approximation_error_pretrained}.
\begin{table*}[h!]
\centering
\caption{We compare the error of first-order approximation from the pretrained initialization and the meta-initialization fine-tuned on all data sources. The results are averaged over $50$ random task subsets. We sample subsets of subsets of size $19$ on Alpaca.}\label{tab_compare_approximation_error_pretrained}
\resizebox{1.8\columnwidth}{!}
{\small\begin{tabular}{@{}cccccccccc@{}}
\toprule
\textbf{Distance}$\backslash$\textbf{RRSS} &  \multicolumn{2}{c}{\textbf{GPT-Neo-1.3B}} & \multicolumn{2}{c}{\textbf{TinyLlama-1.1B}}  \\ \midrule
Initialization $\theta^\star$ &  Pretrained LM &  Fine-tuned on all tasks &  Pretrained LM &  Fine-tuned on all tasks  \\
\midrule
0.05\% & $6_{\pm 0.6}\times10^{-4}$ & $6_{\pm 0.5}\times10^{-5}$  & $8_{\pm 1.1}\times10^{-4}$ & $3_{\pm 0.4}\times10^{-5}$\\
0.10\% & $1_{\pm 0.2}\times10^{-3}$ & $3_{\pm 0.3}\times10^{-4}$  & $2_{\pm 0.4}\times10^{-3}$ & $1_{\pm 0.3}\times10^{-4}$\\
0.15\% & $2_{\pm 0.1}\times10^{-3}$ & $6_{\pm 0.6}\times10^{-4}$  & $3_{\pm 0.4}\times10^{-4}$ & $3_{\pm 0.7}\times10^{-4}$ \\
0.20\% & $3_{\pm 0.3}\times10^{-3}$ &  $1_{\pm 0.2}\times10^{-3}$  & $3_{\pm 0.7}\times10^{-4}$ & $4_{\pm 0.9}\times10^{-4}$ \\
0.25\% & $4_{\pm 0.2}\times10^{-2}$ & $5_{\pm 0.8}\times10^{-3}$  & $7_{\pm 0.6}\times10^{-2}$ & $5_{\pm 0.8}\times10^{-3}$\\ 
\bottomrule
\end{tabular}}
\end{table*}

While using the second-order approximation can further reduce estimation error, it requires computing Hessian-gradient products, which are computationally expensive. Thus, we focus on the first-order approximation in this paper. 

\paragraph{Dimension reduction.} Recall that the dimension of $\nabla h_{\theta^\star}(s, y)$ is the same as the number of trainable parameters in a neural network. Thus, we project the gradients to a much lower dimension using random projection.
Let $P$ be a $p$ by $d$ Gaussian random matrix, whose entries are independently sampled from a Gaussian distribution $N(0, d^{-1})$.
We project the gradients from dimension $p$ to dimension $d$ as
$\tilde g_i = P^{\top} \nabla h_{\theta^\star}(s, y)$.
Then, we solve the logistic regression in dimension $d$. Denote the solution as $\hat \theta_d$.
We set $\hat \theta_S \leftarrow P \hat \theta_d + \theta^{\star}$ to map the projected solution back to $p$-dimensions.

\paragraph{Extension to multi-classification.} To accommodate multi-classification tasks, we can view the cross-entropy loss in the same form as a logistic loss by transforming the model output function. Specifically, for a training example $(s, y)$ where $y$ is a multi-classification label, we can define the model output function as $h_{\theta}(s, y):= \log \big (\frac{p(y|s; \theta)}{1 - p(y|s; \theta)} \big )$ where $p(y|s; \theta)$ is the softmax probability assigned to the correct class. Then, the cross-entropy can be rewritten as the logistic loss as $\ell(s, y; \theta) = - \log p(y|s; \theta).$

\paragraph{Extension to generative tasks.} To accommodate generative tasks, we can view the loss at each output position as a multi-classification loss and average the gradient over every output position. Specifically, for each training example $(s, y)$, suppose that $y$ is a sequence of length $L$ denoted as $y = (y_1, y_2, \ldots, y_L)$. The loss can be written as %
the sum over $L$ conditional probabilities.
We can view this as $L$ multi-class classification losses and apply the above transformation to each position $i = 1, 2, \ldots, L$. Then, we can compute the averaged gradient over the $L$ output positions.

\begin{table*}[t!]
\centering
\caption{Detailed statistics about eight datasets used in the experiments, including the number of tasks, and the link to their data source.}\label{tab_dataset_statistics}
\resizebox{\textwidth}{!}
{\small
\begin{tabular}{@{}llccl@{}}
\toprule
Dataset & \# Tasks & Average size & Category & Source  \\ \midrule
GLUE & 9 & 106,416 & 3 task categories & \href{https://gluebenchmark.com/}{gluebenchmark.com/} \\
FLAN v2 & 1,691 & 59 & 150 task categories & \href{https://huggingface.co/datasets/philschmid/flanv2}{huggingface.co/datasets/philschmid/flanv2}\\
COT & 18 & 8,302 & Chain-of-thought reasoning & \href{https://huggingface.co/datasets/QingyiSi/Alpaca-CoT}{huggingface.co/datasets/QingyiSi/Alpaca-CoT}\\
Alpaca & 38 & 1,073 & Text generation & \href{https://huggingface.co/datasets/tatsu-lab/alpaca}{huggingface.co/datasets/tatsu-lab/alpaca}\\
ToxiGen & 1 & 7,000 & Text generation & \href{https://huggingface.co/datasets/toxigen/toxigen-data}{huggingface.co/datasets/toxigen/toxigen-data} \\
TruthfulQA & 1 & 818 & \makecell{Open-domain QA} & \href{https://huggingface.co/datasets/truthfulqa/truthful_qa}{huggingface.co/datasets/truthfulqa/truthful\_qa}\\
CQA & 100 & 97 & \makecell{Commonsense reasoning} & \href{https://huggingface.co/datasets/tau/commonsense_qa}{huggingface.co/datasets/tau/commonsense\_qa} \\
StrategyQA & 100 & 128 & \makecell{Commonsense reasoning} & \href{https://github.com/eladsegal/strategyqa}{github.com/eladsegal/strategyqa}\\
\bottomrule
\end{tabular}
}
\end{table*}

\begin{table*}[h!]
\centering
\caption{List of hyperparameters for each dataset tested in the experiments, corresponding to the results that we reported in Section \ref{sec_exp}.}\label{tab_hyper_params}
\resizebox{\textwidth}{!}
{\small
\begin{tabular}{@{}lccccccc@{}}
\toprule
Dataset &  Model & Step size & Batch size & Epochs & LoRA rank & Results \\ \midrule
GLUE & BERT-Base, RoBERTa-Base & $5e^{-5}$ & 16 & 5 & Full model & Table \ref{tab_compare_approximation_error} \\ \midrule
\multirow{5}{*}{Alpaca} & Pythia-70M, GPT-2 & $5e^{-5}$ & 16 & 10 & Full model & \multirow{5}{*}{Table \ref{tab_compare_approximation_error} and \ref{table_compare_distance}} \\ 
 & \makecell{FLAN-T5-Base, BloomZ-560M \\ TinyLlama-1.1B, GPT-Neo-1.3B, \\ OPT-1.3B, Gemma-2-2B \\Mistral-7B, Llama-3-8B} & $5e^{-5}$ & 16 & 10 & 16\\ \midrule 
\multirow{3}{*}{StrategyQA} & GPT-2 & $3e^{-4}$ & 8 & 20 & Full model &  \multirow{3}{*}{Table \ref{table_compare_distance}} \\ 
     & \makecell{FLAN-T5-Base, TinyLlama-1.1B,\\ GPT-Neo-1.3B, Llama-3-8B} & $3e^{-4}$ & 8 & 20 & 16\\   \midrule
ToxiGen & TinyLlama-1.1B & $2e^{-5}$ & 128 & 10 & 128 & \multirow{2}{*}{Figure \ref{fig_compare_task_selection}} \\
TruthfulQA & TinyLlama-1.1B & $2e^{-5}$ & 128 & 10 & 128   \\ \midrule
StrategyQA & FLAN-T5-Base & $3e^{-4}$ & 8 & 20 & 16 & \multirow{2}{*}{Figure \ref{fig_compare_runtime_performance_reasoning}}\\ 
CQA & FLAN-T5-Base & $3e^{-4}$ & 8 & 20 & 16 \\ 
\bottomrule
\end{tabular}
}
\end{table*}

\paragraph{Analysis of running time complexity.} As outlined in Table \ref{tab_related_works_comparison}, we compare the runtime complexity of our algorithm with that of the subset selection baselines. Below, we detail the exact number of forward passes required by each method. We denote the number of tasks as $n$. Note that the following number of forward passes is multiplied by the average number of forward passes to train on one task.

    Forward selection: This algorithm employs a greedy subset selection approach. It starts with an empty set and iterates up to $n$ times to select the optimal task to combine with the current subset. At the $i$-th iteration, it fine-tunes one model on the current subset combined with each of the remaining $n-i+1$ unselected tasks. Consequently, the total number of forward passes is at most $\sum_{i=1}^n (n-i+1)\cdot i = \frac{1}{6} n(n+1)(n+2)$. 
    
    Random ensemble: This algorithm fine-tunes models on randomly sampled subsets of tasks and averages their performance on each task to generate a score. It requires $O(n\log n)$ subsets for these scores to converge. In practice, we observe that sampling approximately $10n$ subsets is usually sufficient. If we denote the size of each subset as $\alpha$, the total number of forward passes required is $\alpha n \log n$.
    
    Data selection via important resampling \cite{xie2023data}: This algorithm assesses the $n$-gram features of every data sample without requiring model forward passes. It trains a model on a selected subset of tasks. The total number of forward passes is at most $n$.
    
    Low-rank gradient similarity search \cite{xia2024less}: This algorithm first trains a model on all given tasks. Then, compute the similarity score between feature representations (or gradients) of training and test samples. Lastly, select tasks based on the scores and train a model on the selected subset of tasks. This process requires a total of $3n$ forward passes.
    
    Gradient estimation (\algo{}): This algorithm requires a comparable number of forward passes as DEFT. First, we train a model on all task data and project the gradients of each data sample. Then, instead of computing gradient similarities, we estimate model fine-tuning performances by solving logistic regression on the projected gradients. Finally, we select a subset of tasks based on these estimated performances and train a model on the selected subset. This also requires a total of $3n$ forward passes.

\section{Omitted Experiments from Section \ref{sec_exp}}\label{sec_experiment_details}

We describe each dataset in Table \ref{tab_dataset_statistics}. Among them, FLAN v2 includes a variety of tasks, as it combines four prior instruction tuning datasets. We refer the reader to the paper for their task categories. We used a sampled version of FLAN v2. We partition Alpaca by their instruction types into tasks. %

We experiment with the following models: \href{https://huggingface.co/google-bert/bert-base-uncased}{BERT-Base}, \href{https://huggingface.co/FacebookAI/roberta-base}{RoBERTa-Base}, \href{https://huggingface.co/google/flan-t5-base}{FLAN-T5-Base}, \href{https://huggingface.co/EleutherAI/pythia-70m}{Pythia-70M}, \href{https://huggingface.co/openai-community/gpt2}{GPT-2}, \href{https://huggingface.co/bigscience/bloomz-560m}{BloomZ-560M}, \href{https://huggingface.co/TinyLlama/TinyLlama-1.1B-intermediate-step-1431k-3T}{TinyLlama-1.1B}, \href{https://huggingface.co/facebook/opt-1.3b}{OPT-1.3B}, \href{https://huggingface.co/EleutherAI/gpt-neo-1.3B}{GPT-Neo-1.3B}, \href{https://huggingface.co/google/gemma-2b}{Gemma-2-2B}, \href{https://huggingface.co/mistralai/Mistral-7B-v0.3}{Mistral-7B}, and \href{https://llama.meta.com/llama3/}{Llama-3-8B}.%

In our experiments, we fine-tune the models using the AdamW optimizer. We fine-tune the entire model for smaller models, including BERT-Base, RoBERTa-Base, Pythia-70M, and GPT-2. For other models, we use LoRA for fine-tuning. The training hyperparameters used for each experiment are described in Table \ref{tab_hyper_params}. 

For ToxiGen, we measure the percentage of toxic generations of the model. First, the model is prompted to produce toxic language using human-designed hateful prompts; then, the generations are classified by a toxic content detection model. 
For TruthfulQA, we evaluate the model's accuracy in detecting truthful statements. Given a question and five answer choices, we measure the accuracy of the model in assigning the highest probability to the correct answer.

\subsection{Runtime-accuracy tradeoff}\label{sec_additional_exp_results}

In Table \ref{tab_full_comparison_results}, we report the complete results corresponding to the illustrations shown in the trade-off in Figure \ref{fig_ill_tradeoff} of Section \ref{sec_exp}.

We also compare the GPU hours of each method on StrategyQA. We observe that \algofs{} uses $10.6$ GPU hours and \algore{} uses $14.7$ GPU hours. Corroborating with the FLOPs comparison results, our algorithm reduces the GPU hours required by classic selection methods by $10 \times$. Moreover, our algorithm uses comparable GPU hours to existing selection baselines, with LESS and DEFT taking $10.2$ GPU hours.

We also note that the comparison has a similar trend in the instruction tuning setting. 
\begin{table*}[t!]
\centering
\caption{Performance scores and computation cost (FLOPs) on four datasets, including ToxiGen, TruthfulQA, CommonsenseQA, and StrategyQA. The ToxiGen and TruthfulQA datasets are used to evaluate the instruction tuning setting. The CommonsenseQA and StrategyQA datasets are used to evaluate the chain-of-thoughts fine-tuning setting.
We report the averaged results and standard deviations over three random seeds for each method below.
}\label{tab_full_comparison_results}
{\small
\begin{tabular}{@{}lcccccccc@{}}
\toprule
Dataset   & ToxiGen & TruthfulQA & CommonsenseQA & StrategyQA \\ \midrule
\# all training samples & 272,770 & 272,770 & 9,741 & 12,824 \\
\# test samples  & 7,000 & 818 & 1,221 & 687 \\
Model & TinyLlama-1.1B & TinyLlama-1.1B  & FLAN-T5-Base & FLAN-T5-Base \\ 
\midrule
Metrics (\%) & Toxic generations ($\downarrow$) & Accuracy ($\uparrow$) & Accuracy ($\uparrow$) & Accuracy ($\uparrow$) \\ \midrule
Testing the pretrained model & 70.14 $\pm$ 0.00 & 22.03 $\pm$ 0.00 & 72.23 $\pm$ 0.00 & 51.23 $\pm$ 0.00 \\
Feature transfer & 73.20 $\pm$ 0.44 & 21.03 $\pm$ 0.28 & 51.33 $\pm$ 0.44 & 44.68 $\pm$ 0.45 \\
MTL & 69.44 $\pm$ 0.43 & 23.86 $\pm$ 0.78 & 59.05 $\pm$ 0.69 & 60.84 $\pm$ 0.80 \\
DSIR & 68.10 $\pm$ 0.32 & 24.02 $\pm$ 0.43 & 64.56 $\pm$ 0.28  & 61.49 $\pm$ 0.19\\
DEFT & 65.71 $\pm$ 0.32 & 24.38 $\pm$ 0.61 & 64.78 $\pm$ 0.93 & 61.79 $\pm$ 0.32\\
LESS & 63.90 $\pm$ 0.27 & 24.53 $\pm$ 0.48 & 64.92 $\pm$ 0.32 & 61.93 $\pm$ 0.21\\
Forward selection & 58.50 $\pm$ 0.79  & 27.72 $\pm$ 0.21 & 67.27  $\pm$ 0.68 & 63.54 $\pm$ 0.13\\
Random ensemble & 57.88 $\pm$ 0.28 & 28.21  $\pm$ 0.60 &  67.78 $\pm$ 0.68 & 64.28 $\pm$ 0.24\\ \midrule
\algofs{} & 59.36 $\pm$ 0.53 & 27.25 $\pm$ 0.42 & 67.09 $\pm$ 0.53 & 63.30 $\pm$ 0.50\\
\algore{} & 58.29 $\pm$ 0.66 & 27.93 $\pm$ 0.42 & 67.50 $\pm$ 0.94 & 63.90 $\pm$ 0.43 \\\midrule
\# FLOPs ($\downarrow$) & ToxiGen & TruthfulQA & CommonsenseQA & StrategyQA \\ \midrule
MTL &  5.80 $\times 10^{17}$ & 5.80 $\times 10^{17}$ & 1.19 $\times 10^{16}$ & 1.57 $\times 10^{16}$ \\
DSIR & 8.98 $\times 10^{17}$ & 8.98 $\times 10^{17}$ & 1.78 $\times 10^{16}$ & 2.43 $\times 10^{16}$ \\
DEFT & 1.04 $\times 10^{18}$ & 1.04 $\times 10^{18}$ & 1.84 $\times 10^{16}$ & 2.50 $\times 10^{16}$ \\
LESS & 1.04 $\times 10^{18}$ & 1.04 $\times 10^{18}$ & 1.84 $\times 10^{16}$ & 2.50 $\times 10^{16}$ \\
Forward selection &  5.54 $\times 10^{20}$ & 5.54 $\times 10^{20}$ & 5.98 $\times 10^{17}$ &  8.62 $\times 10^{17}$ \\
Random ensemble & 5.04 $\times 10^{21}$ & 5.04 $\times 10^{21}$ & 5.98 $\times 10^{18}$ & 
7.84 $\times 10^{18}$ \\ \midrule
\algofs{} & 2.81 $\times 10^{18}$ & 2.61 $\times 10^{18}$ & 2.03 $\times 10^{16}$ & 2.62 $\times 10^{16}$  \\
\algore{} & 1.87 $\times 10^{19}$ & 1.53 $\times 10^{19}$ & 3.71 $\times 10^{16}$ & 3.47 $\times 10^{16}$ \\
\bottomrule
\end{tabular}
}
\end{table*}

\subsection{Ablation studies of selection algorithms}\label{sec_ablation}

\textbf{Projection dimension: } Recall that in our algorithm, we project gradients to a lower dimension before solving the logistic regression in the estimation stage. We note that a small value of the projection dimension is sufficient to achieve the approximation results. We vary the projection dimension $d$ between 50, 100, 200, and 400 for running \algo{} on FLAN-T5-Base. The results are shown in Table \ref{table_vary_projection_dimension}. Once $d$ increases above $100$, the error stabilizes around {0.03\%}, so we set $d$ to $100$ by default. $d$ is approximately $7\log(p)$ where $p = 2654208$ is the number of trainable parameters in FLAN-T5-Base.
\begin{table}[h!]
\centering
\caption{Fine-tuning distance vs. the rate of speedup for various values of dimension $d$, computed using the Alpaca and StrategyQA datasets. For both settings, we find that setting $d = 100$ yields the best results. Hence, we use $d = 100$ in our experiments for the random projection.}\label{table_vary_projection_dimension}
\resizebox{0.49\textwidth}{!}
{\begin{tabular}{@{}cccccccc@{}}
\toprule
Alpaca  & \multicolumn{2}{c}{\algofs{}} &  \multicolumn{2}{c}{\algore{}}  \\ \midrule
$\bm{d}$ & \textbf{Distance} & \textbf{Speedup}  & \textbf{Distance} & \textbf{Speedup} \\ \midrule
50  & $4.8 \times 10^{-4}$ & 17.6$\times$ &  $5.2 \times 10^{-4}$ & 43.3$\times$ \\
\textbf{100} & $3.2 \times 10^{-4}$ & 17.6$\times$ &  $3.4 \times 10^{-4}$ & 43.3$\times$ \\
200 & $2.9 \times 10^{-4}$ & 17.5$\times$ & $3.0  \times 10^{-4}$ & 43.2$\times$ \\
400 & $2.8 \times 10^{-4}$ & 17.5$\times$ & $2.9  \times 10^{-4}$ & 43.2$\times$ \\
\midrule
StrategyQA  & \multicolumn{2}{c}{\algofs{}} &  \multicolumn{2}{c}{\algore{}} \\ \midrule
$\bm{d}$ & \textbf{Distance} & \textbf{Speedup} & \textbf{Distance} & \textbf{Speedup} \\ \midrule
50  & $4.6 \times 10^{-4}$   & 30.5$\times$ & $5.6 \times 10^{-4}$ & 44.9$\times$ \\
\textbf{100} & $3.0 \times 10^{-4}$  & 30.5$\times$ & $3.8 \times 10^{-4}$ & 44.9$\times$\\
200 & $2.8  \times 10^{-4}$ & 30.4$\times$ &$3.6  \times 10^{-4}$ & 44.8$\times$\\
400 & $2.7  \times 10^{-4}$ & 30.4$\times$ &$3.4  \times 10^{-4}$ & 44.8$\times$\\
\bottomrule
\end{tabular}}
\end{table}

\paragraph{Number of random subsets $m$ in the ensemble method.} Recall that in applying  \algore{}, the score in the random ensemble for each task is estimated by averaging the value functions of $m$ subsets that contain the task.  
For the strategy QA data with $n = 100$ datasets,  we observe that the estimated higher-order task affinity converges using $m = 1000$ subsets. We computed the distance between the estimated scores $T$ using $m$ subsets and the final scores $T^*$, observing that the distance sharply decreases as $m$ increases, eventually converging to zero. This analysis was conducted for different subset sizes, with $\alpha = 0.25n$, $\alpha = 0.5n$, and $\alpha = 0.75n$, respectively. The distance stabilizes as $m$ approaches 1000, indicating that the estimated task affinity scores converge.

One might hypothesize that using 1000 or more subsets achieves comparable performance. To test this, we run the subset selection algorithm using $m  = 500, 1000, 1500, 2000$.  We notice no further performance gain in the downstream test accuracy when using more subsets than $1000$. Thus, we set $m = 1000$ in the random ensemble method.

\paragraph{Choice of subset size in random ensemble.}  We vary subset size $\alpha$ between $0.25n$, $0.5n$, and $0.75n$. We find the $T_i$ scores all converge at a similar rate.  Additionally, using a subset size of $0.75n$ yields better performance in downstream selection.  The reason is that this size is closer to the size of the selected subset (i.e., $0.7n$).

\subsection{Complete subset selection pseudocodes}

We provide the complete procedure for forward search and random ensemble with gradient estimation below.
In particular, both procedures use Algorithm \ref{alg_estimate_performance} as a subroutine.
For further details, see our code implementation online at \url{https://github.com/VirtuosoResearch/Scalable-finetuning}.
\begin{algorithm}[h!]
\caption{{\sc \textbf{GradEx-FS}}}\label{alg_forward_selection}
\raggedright
\textbf{Input}: The training datasets of $n$ data sources\\
\textbf{Require}: A multitask fine-tuning algorithm $f$\\
\textbf{Output}: A subset of selected tasks $S^\star$ \\
\begin{algorithmic}[1]
    \State Initiate an empty set $S_{\text{cur}} = \set{}$
    \For{$i = 1, 2, \ldots, n$}
    \For{each task not selected in current subset $j \in [n]/S_{\text{cur}}$ }
    \State Add task $j$ to current subset: $S_{j} \leftarrow S_{\text{cur}} \bigcup \set{j}$ 
    \State Estimate $\hat f (S_j) $ using \algo{}
    \EndFor
    \State Choose the task with the highest value function $j^{\star} \leftarrow \arg \max_{j \in [n]/S_{\text{cur}}} \hat f(S_{j})$
    \State Add the dataset to the current subset: $S_{\text{cur}} \leftarrow S_{\text{cur}} + j^{\star}$, if $\hat f(S_{j}) < \hat f(S_{\text{cur}})$; otherwise, break
    \EndFor
    \State Return the selected subset $S^\star \leftarrow S_{\text{cur}}$ 
\end{algorithmic}
\end{algorithm}
\begin{algorithm}[h!]
\caption{{\sc \textbf{GradEx-RE}}}\label{alg_random_ensemble}
\raggedright
\raggedright
\textbf{Parameter}: The size of each sampled subset $\alpha$, and the number of sampled subsets $m$; a selection threshold $\gamma$\\
\begin{algorithmic}[1]
    \For{$k = 1, \dots, m$}
    \State Sample a random subset $S_k$ from $\set{1,2,\dots,n}$ with size $\alpha$
    \State Estimate $\hat f(S_k)$ using \algo{}
    \EndFor
    \State Estimate a score for each dataset $i$ as the value function averaged over subsets that include $i$:
    \[
    T_{i} = \frac{1}{n_{i}} \sum_{ 1\leq k \leq m:\ i \in S_k} \hat f\big(S_k\big), \text{ for } 1 \leq i \leq n
    \]
    \vspace{0.01in}
    \State Select a subset of tasks by thresholding the scores: $S^\star = \set{i \ | \  T_i < \gamma, \ \forall \ i = 1, 2, \ldots, n}$
\end{algorithmic}
\end{algorithm}